\documentclass[runningheads]{llncs}

\usepackage{eccv}

\usepackage{eccvabbrv}
\usepackage{amssymb}
\usepackage{graphicx}
\usepackage{booktabs}
\usepackage{caption}
\usepackage[ruled,linesnumbered]{algorithm2e}
\usepackage{bm}
\usepackage[accsupp]{axessibility}  
\usepackage{tabularx}
\newcolumntype{C}{>{\centering\arraybackslash}X}
\newcolumntype{R}{>{\raggedleft\arraybackslash}X}
\newcolumntype{L}{X}

\usepackage{hyperref}
\usepackage{orcidlink}

\begin{document}

\title{Crowd-SAM: SAM as a Smart Annotator for Object Detection in Crowded Scenes} 

\titlerunning{Crowd-SAM: Smart Annotation for Crowded Scenes}

\author{Zhi Cai\inst{1,2}\orcidlink{0009-0003-6321-1286} \and
Yingjie Gao\inst{1,2}\orcidlink{0009-0003-4037-4701} \and
Yaoyan Zheng\inst{1,2}\orcidlink{0009-0002-2591-4020} \\ 
Nan Zhou\inst{1,2}\orcidlink{0000-0003-3443-6171} \and
Di Huang\inst{1,2}\thanks{Corresponding author}\orcidlink{0000-0002-2412-9330}
}

\authorrunning{Z. Cai \emph{et al}.}

\institute{
$^1$SKLSDE, Beihang University, Beijing, China\\
$^2$IRIP Lab, SCSE, Beihang University, Beijing, China\\
\email{\{caizhi97, gaoyingjie, yaoyanzheng, zhounan0431, dhuang\}@buaa.edu.cn}
}

\maketitle

\begin{abstract}
Object detection is an important task that finds its application in a wide range of scenarios. Generally, it requires extensive labels for training, which is quite time-consuming, especially in crowded scenes. 
Recently, Segment Anything Model (SAM) has emerged as a powerful zero-shot segmenter, offering a novel approach to instance segmentation. However, the accuracy and efficiency of SAM and its variants are often compromised when handling objects in crowded scenes where occlusions often appear.
In this paper, we propose Crowd-SAM, a SAM-based framework designed to enhance the performance of SAM in crowded scenes with the cost of few learnable parameters and minimal labeled images.
We introduce an efficient prompt sampler (EPS) and a part-whole discrimination network (PWD-Net), facilitating mask selection and contributing to an improvement in accuracy in crowded scenes. Despite its simplicity, Crowd-SAM rivals state-of-the-art fully-supervised object detection methods on several benchmarks including CrowdHuman and CityPersons. 
Our code is available at \href{https://github.com/FelixCaae/CrowdSAM}{https://github.com/FelixCaae/CrowdSAM}.
\keywords{Detection in crowded scenes \and Few-shot learning \and Segment anything model } 
\end{abstract}

\section{Introduction}
\label{sec:intro}

Object detection in crowded scenes is a fundamental task in areas such as autonomous driving and video surveillance.
The primary focus lies in identifying and locating densely packed common objects like pedestrians and vehicles, where occlusions present significant challenges.
Great progress has been made in recent years, including two-stage methods~\cite{wang2018repulsion,zhang2018occlusion} and query-based methods~\cite{lin2020detr,zheng2022progressive,gao2023selecting}.
However, these methods mainly follow a supervised manner and necessitate extensive labeled training samples, incurring a considerable annotation cost of approximately 42.4 seconds per object~\cite{su2012crowdsourcing} 
The density and complexity of crowded scenes further aggravate the annotation burden. \footnote{On average, an image contains approximately 22 objects in CrowdHuman \cite{shao2018crowdhuman} and 7 in MS COCO \cite{coco}}


The high cost of collecting object annotations drives the exploration of alternatives such as few-shot learning~\cite{wang2020frustratingly,sun2021fsce,qiao2021defrcn}, weakly supervised learning~\cite{tang2017multiple, Zeng_2019_ICCV}, semi-supervised learning~\cite{liu2021unbiased,xu2021end,tang2021humble, Liu_2022_CVPR}, and unsupervised learning~\cite{liu2020self,dai2021up,xie2021detco,bar2022detreg,Li_2023_ICCV}. The best-performing ones, \ie semi-supervised object detection (SSOD), leverage both labeled and unlabeled data for training and achieve a big success on common benchmarks \eg PASCAL VOC~\cite{hoiem2009pascal} and COCO~\cite{coco}. Unfortunately, SSOD introduces extra complexity such as complicated augmentations and online pseudo-labeling.

Recently, prompt-based segmentation models have received increasing attention due to their flexibility and scalability. Particularly, Segment Anything Model (SAM)~\cite{kirillov2023segment} show its high capability to effectively and accurately predict the masks of regions specified by prompts, in any form of points, boxes, masks, or text descriptions. Recognizing its exceptional potential, researchers have made many efforts to adapt it for various vision tasks such as medical image recognition~\cite{ma2024segment}, remote sensing analysis~\cite{chen2024rsprompter,gui2024remote}, industrial defect detection~\cite{ye2024sam}, \etc 

Despite the huge progress~\cite{hqsam,wei2023semantic,zhang2023faster} following SAM, applying SAM for object detection in crowded scenes is seldom studied. In this paper, we investigate the potential of SAM in such cases with two motivations. First, SAM is pre-trained on a very large dataset \ie SA-1B that contains most of the common objects and it is thus reasonable to utilize the knowledge to facilitate labeling massive data and training a brand-new detector. Second, SAM demonstrates a robust segmentation ability in handling complicated scenes characterized by clustered objects that are difficult for an object detector trained from scratch.

\begin{figure}[ht]
    \centering
    \includegraphics[width=1.0\textwidth]{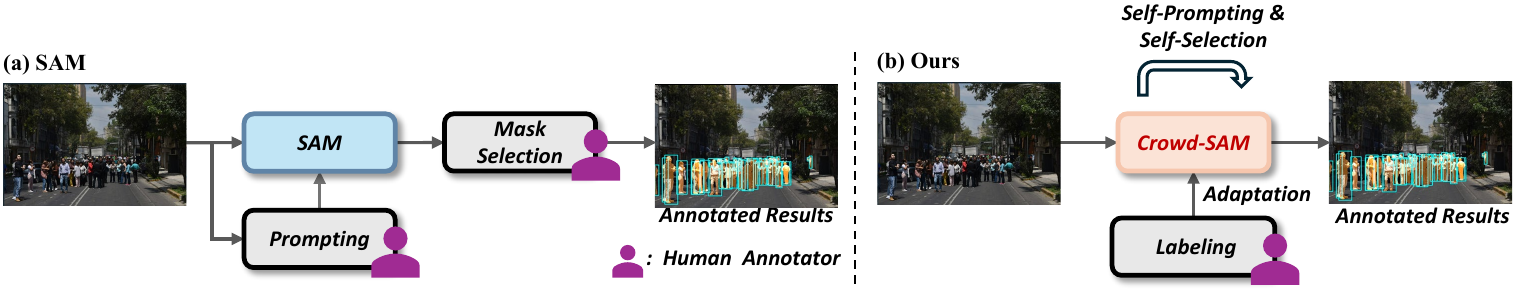}
    \caption{
    Pipeline comparison between SAM and Crowd-SAM. Crowd-SAM only requires a  few labeled images and can automatically recognize target objects.
    }
    \label{fig:compare_work}
\end{figure}

To this end, we propose Crowd-SAM, a smart annotator powered by SAM for object detection in crowded scenes. As depicted in~\cref{fig:compare_work}, we introduce a self-promoting approach based on DINOv2 to alleviate the cost of human prompting. Our method employs dense grids equipped with an Efficient Prompt Sampler (EPS) to cover as many objects as possible at a moderate cost. To distinguish the masks from multiple outputs precisely in occluded scenes, we design a mask selection module, termed Part-Whole Discrimination Network (PWD-Net) that learns to differentiate the output with the highest quality in Intersection over Union (IoU) score. With a lightweight model design and fast training schedule, it delivers considerable performance on public benchmarks including CrowdHuman~\cite{shao2018crowdhuman} and CityPersons~\cite{zhang2017citypersons}. 

Our contributions can be summarized as follows:
\begin{enumerate}
  \item We propose Crowd-SAM, a self-prompted segmentation method, for labeling images containing clustered objects, producing accurate results with only a few examples.
  \item We design two novel components of Crowd-SAM, \ie EPS and PWD-Net, which effectively unleash the ability of SAM on crowded scenes.
  \item We conduct comprehensive experiments on two benchmarks to demonstrate the effectiveness and generalizable nature of Crowd-SAM.
\end{enumerate}

\section{Related Work}
\textbf{Object Detection.} 
General object detection aims to identify and locate objects and is mainly divided into two categories: \ie one-stage detectors and two-stage detectors. One-stage detectors predict bounding boxes and class scores by using image features\cite{redmon2016you,liu2016ssd,lin2017focal}, while two-stage detectors first generate region proposals and then classify and refine them~\cite{faster,girshick2014rich,girshick2015fast}. Recently, end-to-end object detectors \eg DETR\cite{detr,deformabledetr,dino} have replaced the hand-crafted modules such as Non-Maximum Suppression (NMS) by adopting one-to-one matching in the training phase, showing great potential in a wide variety of areas.

However, applying these detectors directly to pedestrian detection tends to incur performance degradation due to the fact that pedestrians are often crowded with occlusions appearing.
Early work~\cite{mao2017can} proposes to integrate extra features into a pedestrian detector to explore low-level visual cues, while follow-up methods \cite{chi2020pedhunter,zhang2019double} attempt to utilize the head areas for better representation learning. In \cite{zhang2019double}, an anchor is associated with two targets, the whole body, and the head part, to achieve a more robust detector from joint training. AdaptiveNMS~\cite{Liu_2019_CVPR} adjusts the NMS threshold by predicting the density of pedestrians. Alternative methods focus on the design of loss functions to improve the training process. For example, RepLoss~\cite{wang2018repulsion} encourages the prediction consistency of the same target while repels the ones from different targets.  Recently, Zheng~\etal \cite{zheng2022progressive} models the relation of queries to improve DETR-based detectors in crowded scenes and achieves remarkable success. 
Although these works have pushed the boundaries of object detection in crowded scenes to a new stage, they all rely on a large number of labeled samples for training, which is labor-intensive. This limitation inspires us to develop label-efficient detectors and automatic annotation tools, with the help of SAM.

\textbf{Few-Shot Object Detection (FSOD).} This task aims to detect objects of novel classes with limited training samples. FSOD methods can be roughly classified into meta-learning based \cite{yan2019meta,kang2019few} and fine-tuning based ones~\cite{wang2020frustratingly,sun2021fsce,qiao2021defrcn}. 
Meta-RCNN~\cite{yan2019meta} processes the query and support images in parallel via a siamese network. The Region of Interest (RoI) features of the query are fused with the class prototypes to effectively transfer knowledge learned from the support set. TFA~\cite{wang2020frustratingly} proposes a simple two-stage fine-tuning method that only fine-tunes the last layers of the network. FSCE~\cite{sun2021fsce} introduces a supervised contrastive loss in the fine-tuning stage to mitigate misclassification issues. De-FRCN~\cite{qiao2021defrcn} stops the gradient from the RPN and scales the gradient from R-CNN\cite{faster}, followed by a prototypical calibration block to refine the classification scores. 

\textbf{Segment Anything Models.} 
SAM~\cite{kirillov2023segment}, a visual foundation model for segmentation tasks, is trained on the SA-1B dataset using a semi-supervised learning paradigm. Its exposure to this vast repository of training samples renders it a highly capable class-agnostic model, effectively handling a wide range of objects in the world. Despite its impressive performance in solving segmentation tasks, it suffers from several issues like domain shift, inefficiency, class-agnostic design, \etc  HQ-SAM~\cite{hqsam} is proposed to improve its segmentation quality by learning a lightweight adapter. Fast-SAM~\cite{zhao2023fast} and Mobile-SAM~\cite{zhang2023faster} focus on fastening the inference speed of SAM by knowledge distillation.
RSprompt~\cite{chen2024rsprompter} enables SAM to generate semantically distinct segmentation results for remote sensing images by generating appropriate prompts. Med-SA~\cite{wu2023medical} presents a space-depth transpose method to adapt 2D SAM to 3D medical images and a hyper-prompting adapter to achieve prompt-conditioned adaptation. Unfortunately, these approaches necessitate a considerable amount of labeled data for effective adaptation, making them impractical for crowded scenes where annotation costs are prohibitive. Different from them, Per-SAM~\cite {zhang2024personalize} and Matcher~\cite{liu2024matcher} teach SAM to recognize specified objects with only one or few instances by extracting training-free similarity priors. SAPNet~\cite{wei2023semantic} builds a weakly-supervised pipeline for instance segmentation.
Although these approaches reduce data requirements, they still lag behind the demands of crowded scenes, such as pedestrian detection, particularly with occlusions.

\begin{figure}[t]
    \centering
    \includegraphics[width=1.0\columnwidth]{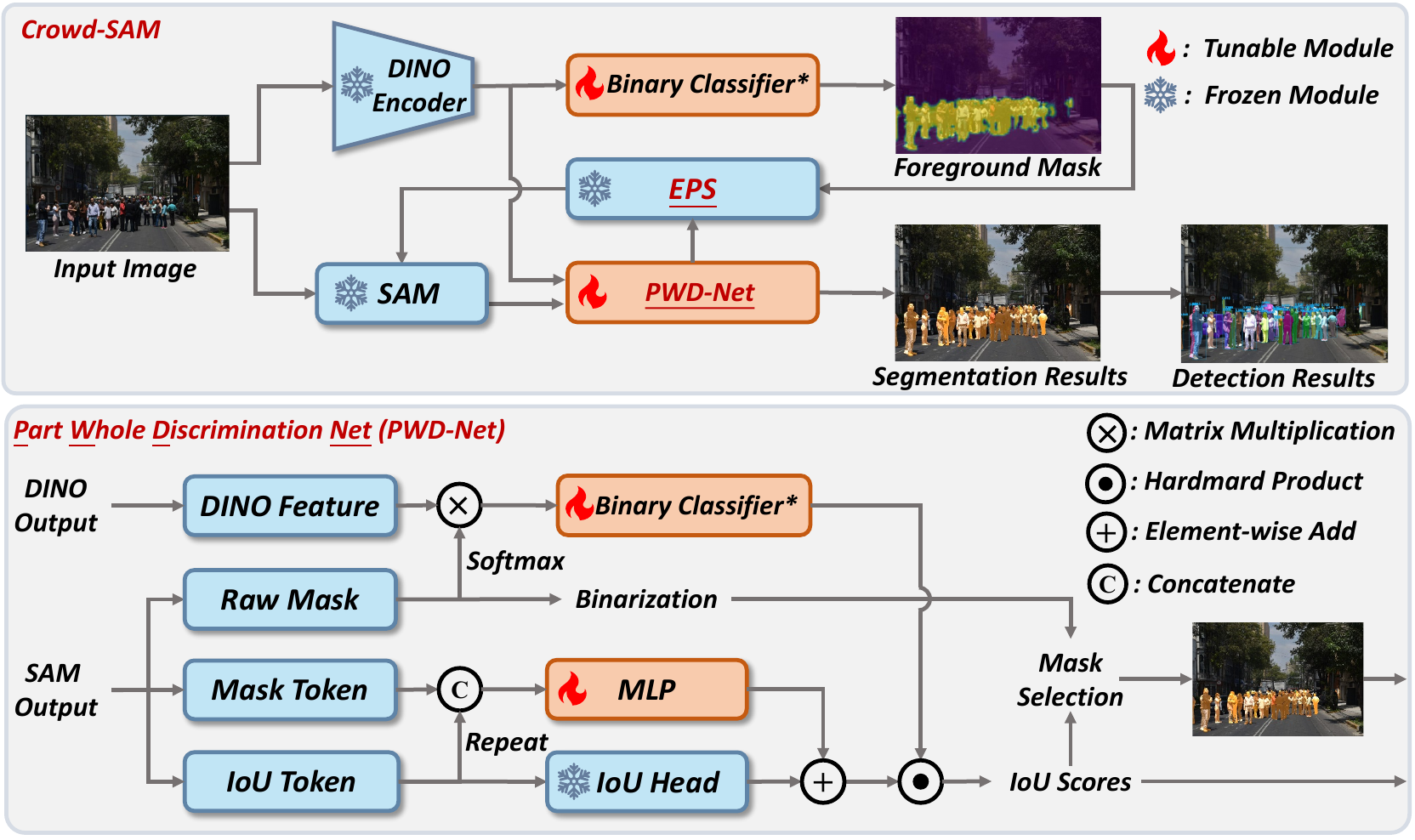}
    \caption{The pipeline of Crowd-SAM shows the interaction between different modules. DINO encoder and SAM are frozen in the training process. * represents the parameters that are shared. For simplicity, the projection adapter of DINO is dismissed.}
    \label{fig:pipeline}
    
\end{figure}

\section{Method}

\subsection{Preliminaries}
\textbf{SAM}~\cite{kirillov2023segment} is a powerful and promising segmentation model that comprises three main components: (\textbf{a}) an image encoder responsible for feature extraction; (\textbf{b}) a prompt encoder designed to encode the geometric prompts provided by users; and (\textbf{c}) a lightweight mask decoder that predicts the masks conditioned on the given prompts. Leveraging extensive training data, SAM demonstrates impressive zero-shot segmentation performance across various benchmarks. In particular, SAM makes use of points and boxes as prompts to specify interested regions.


\textbf{DINO} \cite{caron2021emerging} represents a family of Vision Transformers (ViT) \cite{dosovitskiy2020image} learned in a self-supervised manner designed for general-purpose applications. During its training, DINO employs a self-distillation strategy akin to BYOL \cite{grill2020bootstrap}, fostering the learning of robust representations.
DINOv2~\cite{oquab2023dinov2} strengthens the foundation of DINO by integrating several additional pre-training tasks, improving its scalability and stability, especially for large models \eg ViT-H (1 billion parameters). Thanks to its enhancement, DINOv2 shows a strong representation ability, in particular for the task of semantic segmentation.

\subsection{Problem Definition and Overall Framework}
\textbf{Problem Definition}. As shown in~\cref{fig:compare_work}, our goal is to detect objects (\eg pedestrians) in crowded scenes with few annotated data. We formulate this problem as a one-class few-shot detection task. A common few-shot pipeline is to divide data into the base split and the novel split. Differently, we directly use the data of the target class for model training, as the foundation models have already been trained on massive data. In particular, we employ segmentation masks as intermediate results which can be easily converted to bounding boxes. During the training and evaluation processes, only box annotations are provided.


\textbf{Naive Study on SAM Auto-generator}. The prompt number affects the performance of SAM and we analyze this issue for crowded scenes. In this case, we conduct several naive studies on CrowdHuman~\cite{shao2018crowdhuman} with the auto generator of SAM, which utilizes grid points to search every region. ~\cref{tab:sam_baseline} conveys three key observations: \textbf{(1)} dense grids are necessary for crowded scenes; \textbf{(2)} the ambiguity of distributions of point prompts and class-agnostic prompts incur many false positives (FPs); \textbf{(3)} the decoding time is a non-negligible burden when the grid size is large. Therefore, the \textbf{dense prompts} and \textbf{FP removal} are key aspects in designing SAM-based methods for detection/segmentation tasks in crowded scenes. 

\begin{table}[t]
    \centering
        \caption{Comparison in terms of recall, average FPs, and decoding time ($T$) of different grid sizes ($N_{G}$) adopted by the SAM generator on CrowdHuman~\cite{shao2018crowdhuman}. The oracle model derives the prompts by computing the center of ground truth boxes. The decoding time is collected on a 3090 Ti GPU card.}
    \scalebox{1.0}{
    \begin{tabular}{c|cccc|c}
    \toprule
          $N_{G}$ & 16 & 32 & 64 & 128 & Oracle \\
          \hline
          Recall & 33.6&   58.0 & 63.4 & 76.0 & 91.4\\
          \hline
          avg. FPs & 51 & 112 & 227 & 485 & -\\
          \hline
          $T$ (s) & 0.059& 0.22 & 0.83 & 3.2 & 0.045 \\
    \bottomrule
    \end{tabular}
    }

    \label{tab:sam_baseline}
\end{table}

\textbf{Overall Framework}. Inspired by the studies above, we equip SAM~\cite{kirillov2023segment} with several proper components to achieve an accurate and efficient annotation framework, as illustrated in~\cref{fig:pipeline}. To accurately locate clustered objects, we employ the foundational model DINOv2\cite{oquab2023dinov2} to predict a semantic heat map, a task that can be formulated with a simple binary classifier. To discriminate the output masks that is a mixture of correct masks, backgrounds, and part-level masks, we design Part-Whole Discrimination Network (PWD-Net) that takes as input both the learned tokens from SAM and the semantic-rich tokens from DINOv2 to re-evaluate all the outputs. Finally, to handle the redundancy brought by the use of dense grids, we propose an Efficient Prompt Sampler (EPS) to decode the masks at a moderate cost. 


We introduce the details of our methods in the following sections.

\subsection{Class-specific Prompt Generation\label{sec:prompt_gen}}

Generating a unique point prompt for each object (\eg pedestrian) in crowded scenes is de facto a non-trivial problem. Thus, we take a step back and study how to detect objects with multiple prompts associated with one object and apply the proper post-processing techniques to remove duplicates. To this end, we adopt a heatmap-based prompt generation pipeline that initially classifies the regions and then generates prompts from the positive regions.
 
For the input image $I \in \mathbb{R}^{H \times W \times 3} $, we first use a pre-trained image encoder $E_{D}$ to extract semantic rich features. To better transfer the pre-trained features to object segmentation, we add an MLP block after the final output layer and thus obtain the adapted features $F_{DINO}\in \mathbb{R}^{{H\over s} \times {W\over s} \times C}$, where $s$ is the patch size of DINOv2~\cite{oquab2023dinov2} and $C$ is the output channel. Then, we employ a segmentation head $\text{Head}_{CLS}$ to classify $F_{DINO}$ pixel by pixel, resulting in a heatmap $\hat{\mathbb{H}} \in [0,1]^{{H\over s} \times {W\over s}}$ that indicates the locations of objects as $\hat{\mathbb{H}} = \text{Head}_{CLS}(F_{DINO})$.

Given only the bounding box annotations $B\in [0,1]^{N_{B} \times 4}$, where $N_{B}$ is the number of targets, this binary segmentation head can be optimized with box-level supervision. However, the coarse boundaries tend to incur considerable points scattering in background regions. To alleviate this issue, we use SAM~\cite{kirillov2023segment} to generate high-quality mask-level pseudo labels, which is illustrated in~\cref{fig:qualitve_comparison}, with ground truth (GT). The decoded masks are then merged into a single foreground mask $\mathbb{H}\in \{0,1\}^{256 \times 256}$. We use the dice loss for training the adapter and segmentation head with the generated pseudo masks as follows:

\begin{equation}
{\cal L}_{fg} = dice(f(\hat{\mathbb{H}}), \mathbb{H}),
    \label{eq:bin_loss}
\end{equation}
where $f$ is an up-sampling function that resizes $\mathbb{H}$ to $256\times 256$.
During inference, we add a threshold $t$ for mask binarization, which is simply set at 0.5 in our experiments. 
The binarized masks are mapped to point prompts $P_{G}$ which only contain those in positive regions. 

\subsection{Semantic-guided Mask Prediction}
\begin{figure}
    \centering
    \includegraphics[width=1\columnwidth]{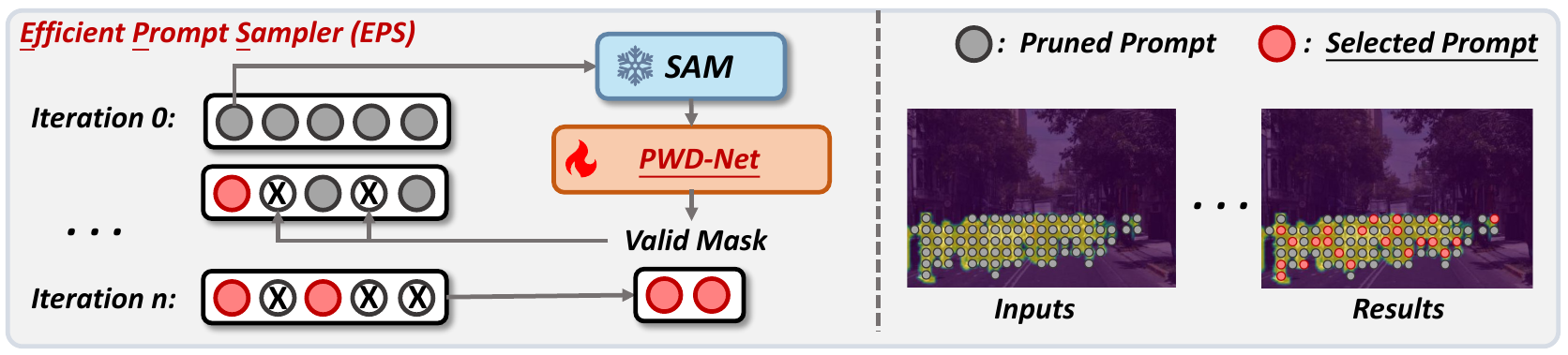}
        \caption{Illustration of EPS. PWD-Net produces valid masks with a threshold.  In each iteration, we prune prompts (\emph{with a cross above}) that fall inside valid masks .}
    \label{fig:eps}
\end{figure}

Given the proposals generated in~\cref{sec:prompt_gen}, our further aim is to efficiently decode the dense prompts and accurately discriminate the generated masks. As depicted in~\cref{fig:eps}, each instance contains a set of prompts due to the density of grids. Supposing that only one in-position prompt is required for mask prediction in SAM, decoding all the prompts would lead to not only a waste of computation but also more FPs for some poorly located prompts.

\textbf{Efficient Prompt Sampler (EPS).} To address this challenge, we introduce EPS, which dynamically prunes prompts based on the confidence of decoded masks. This method is elaborated in~\cref{alg:eps}. Beginning with the generated point prompts list $P_{G}$, EPS, in each iteration, samples a batch of prompts $P_{B}$ from $P_{G}$ using uniform random sampling. The sampled prompts $P_B$ are then appended to the output points list $P_S$. Subsequently, we employ the SAM generator with the batched prompts $P_{B}$ to produce masks $M$. Furthermore, we aggregate discriminant confidence scores $S$ from PWD-Net, which are employed to select valid masks $M'$ using a score threshold $T$, by using $M'=M[S>T]$. Refer to a detailed exposition of PWD-Net in the subsequent section. The valid masks represent regions that are believed to be well segmented, and we thus remove the points that have already been covered by any mask $m$ in $M'$ from the prompts list $P_G$. The iteration halts when $P_G$ is empty. Additionally, EPS establishes a stopping criterion with a parameter $K$, terminating the sampling process once the total sample count reaches this limit. This parameter is instrumental in managing the overall decoding cost. In our experiments, $K$ is empirically set to 500 to strike a balance between efficiency and recall. 




\begin{algorithm}
	\caption{Process of Efficient Prompt Sampler (EPS).}
	\label{alg:eps}
	\KwIn{
feature extracted from image $I$ by image encoder of SAM: $F_{SAM}$; 
generated prompt list: $P_G=\{p_1,p_2,...,p_N\}$; 
SAM generator: $G$; 
mask confidence threshold: $T$; 
target sampled prompt list size: $K$ 
}
        \KwOut{sampled prompt list: $P_S$; valid mask list: $M_S$}  
	\BlankLine
	$P_S\leftarrow \emptyset$;
	
	\While{\textnormal{$|P_G|>0$ and $|P_S|<K$}}{
		Sample a batch of prompts $P_B \subseteq P_G$ with uniform random sampling;

        $P_S\leftarrow P_S\cup P_B$;

        $P_G\leftarrow P_G\setminus P_B$;
  
        Generate masks $M$ corresponding to $P_B$ by $G(F_{SAM}, P_B)$;

        Select valid masks $M'$ according to $T$ from $M$;
        
        $M_S\leftarrow M_S\cup M'$;
        
        \For{$p \in P_G$}{
            \If{\textnormal{$\exists m \in M' \, \text{such that} \, p\in m$}}{
                $P_G\gets P_G\setminus \{p\}$
            }
        }
	}
\end{algorithm}

\textbf{Part-Whole Discrimination Network (PWD-Net).} Given the raw masks predicted by SAM conditioned on the sampled batch of prompts $P_{B}$, we design an automatic selector for choosing the best-fitting mask. It is expected to optimize the outputs in two aspects: \textbf{(i)} refining the output IoU score according to the quality of related masks if they are positive samples and \textbf{(ii)} suppressing the scores of samples that fall in background regions. 

Illustrated in~\cref{fig:pipeline}, for the masks generated corresponding to $N$ prompts, we leverage the \textit{Mask Tokens} and \textit{IoU Tokens} within the mask decoder of SAM along with the sophisticated features extracted by the self-supervised pre-trained model DINOv2~\cite{oquab2023dinov2}. ${\cal M}$ and ${\cal U}$ are responsible for mask decoding and IoU prediction in the SAM Mask Decoder, respectively. Thus, we suppose that they contain shape-aware information, which is helpful in discriminating the mask. These components enable us to compute a discriminant confidence score $S$ for each specific prompt in a few-shot scenario. Initially, the refined IoU score $S_{iou}$ is computed as follows:
\begin{align}
S_{iou}=\text{Head}_{par}(\text{Concat}(\text{Repeat}({\cal U}), {\cal M}))+\text{Head}_{IoU}({\cal U}),
\label{eq:s_iou}
\end{align}
where $\text{Head}_{par}$ is a parallel IoU head, consisting of an MLP block; ${\cal U}\in \mathbb{R}^{N\times 1\times C}$ and ${\cal M}\in \mathbb{R}^{N\times 4\times C}$ denote the \textit{IoU Tokens} and \textit{Mask Tokens} respectively. Notice in~\cref{eq:s_iou}, $S_{iou}$ is a sum of outputs from two individual heads, the parallel adapter head $\text{Head}_{par}$ and the original $\text{Head}_{IoU}$. We freeze the parameters of $\text{Head}_{IoU}$ to avoid overfitting in few-shot learning. Moreover, since these two tokens, ${\cal M}$ and ${\cal U}$ have different shapes, we repeat ${\cal U}$ by 4 times and concatenate these two tokens in the channel dimension. The refined score $S_{iou}\in \mathbb{R}^{N\times 4}$ encapsulates the quality of the generated masks by assessing the texture-aware feature from ${\cal M}$ and ${\cal U}$.

Further, by harnessing the semantic feature embedded in the self-supervised pre-trained model DINOv2, along with the mask data $\hat{M}$, we calculate the discriminant score $S_{cls}$ :
\begin{align}
S_{cls} = \sigma({\text{Head}_{CLS}({\cal O})}), {\cal O} = \text{Pool}(d(\text{Softmax}(\hat{M})) \circ F_{DINO}).
\end{align}
Here, ${\cal O}\in \mathbb{R}^{N\times C}$ denotes the extracted \textit{Semantic Token}, and $\hat{M}\in \mathbb{R}^{N\times H\times W\times 4}$ and $F_{DINO}\in\mathbb{R}^{N\times h\times w\times C}$ represent the masks generated by SAM\cite{ma2024segment} and the features extracted by DINOv2, respectively. We denote $d$ as a down-scale function that resizes the mask to $h\times w$, consistent with $F_{DINO}$. Pool is a global pooling function that conducts mean pooling on the x-axis and y-axis and $\circ$ is the Hadamard product. The discriminant score $S_{cls}\in \mathbb{R}^{N\times4}$ predicts whether masks belong to the foreground or background. $\text{Head}_{CLS}$ shares the same parameters with the binary classifier introduced in~\cref{sec:prompt_gen}. Finally, we calculate a joint score of discrimination and estimate the quality for masks by simply multiplying the two scores: $S=S_{iou}\cdot S_{cls}$. 

During training, prompts sampled from real masks are taken as input. Regarding the prompts within the foreground, the discriminant confidence score aims to accurately predict the IoU of the generated and real masks. Conversely, for the prompts within the background, the score is ideally $0$. Hence, the loss function for this aspect is formulated as follows:
\begin{align}
s^i_{target} =  
\begin{cases} 
    IoU(m^i,m^i_{GT}), & m^i_{GT} \in M^{bg}_{GT}, \\  
    0, & m^i_{GT} \in M^{fg}_{GT} ,
\end{cases}
\end{align}
\begin{align}
{\cal L}_{iou}=\text{MSE}(S,S_{target})&.\label{eq:iou_loss}
\end{align}
Here, $s^i_{target}$ denotes the target score of the mask $m^i$ generated by the $i^{th}$ prompt, and $S_{target}=\{s^i_{target}\}_{i=1}^N;M=\{m^i\}_{i=1}^N;M_{GT}=\{m^i_{GT}\}_{i=1}^N=M_{GT}^{bg}\cup M_{GT}^{fg}$. 

\subsection{Training and Inference}
The total training loss of the entire framework combines~\cref{eq:bin_loss} and~\cref{eq:iou_loss}:
\begin{equation}
    {\cal L} = {\cal L}_{fg}  + {\cal L}_{iou}  
\end{equation}

At inference, we select the mask with the highest confidence score $S$ among the $4$ masks as the output of PWD-Net, which we denote as $M\in\mathbb{R}^{N\times H\times W}$ and $S_{o}\in\mathbb{R}^{N}$. We adopt a window cropping strategy to enhance the performance on small objects as \cite{kirillov2023segment}. This strategy slices the whole image into overlapping crops where each crop is individually processed. The final results are merged from the outputs of each crop and here we apply NMS to remove duplicate proposals.

\section{Experiments}

\begin{table}[ht]
    \centering
        \caption{Comparative results (\%) on CrowdHuman~\cite{shao2018crowdhuman} \emph{val}. All the SAM-based methods adopt ViT-L~\cite{dosovitskiy2020image} as the pre-trained backbone (SRCNN denotes Sparse R-CNN~\cite{sparsercnn}, a baseline in \cite{zheng2022progressive} and * represents using the multi-crop trick).} 
    \scalebox{1.0}{
    \begin{tabular}{lccccccc}
    \toprule
    Method & Backbone & \#Shot & AP  & MR$^{-2}$ & Recall & Secs/Img\\
    \hline
    \multicolumn{3}{l}{\emph{Fully supervised object detectors} }\\
    \hline
    ATSS~\cite{atss}  &ResNet-50~\cite{he2016deep}& Full & 80.3 & 59.7 & 86.1  & 0.051\\
    FCOS~\cite{tian2019fcos}  &ResNet-50~\cite{he2016deep}& Full & 76.3 & 65.5 & 82.6 & 0.045\\
    Iter-SRCNN~\cite{zheng2022progressive}  &ResNet-50~\cite{he2016deep}& Full & 85.9 & 58.3 &93.3& 0.25 \\
    DINO~\cite{dino} &ResNet-50\cite{he2016deep} &Full & 86.7 & 57.6 & 94.5 & 0.27\\
    \hline
    \multicolumn{3}{l}{\emph{Few-shot  object detectors} }\\
    \hline
    TFA~\cite{wang2020frustratingly} & ResNet-101~\cite{he2016deep} & 10 & 46.9 & 84.3 & 57.9 & 0.067\\
    FSCE~\cite{sun2021fsce} & ResNet-101~\cite{he2016deep} &10 &43.0 & 84.7 & 50.0 & 0.072\\
    De-FRCN~\cite{qiao2021defrcn}& ResNet-101~\cite{he2016deep} &10 & 46.4 & 85.9 & 65.5 & 0.072\\
    \hline
    \multicolumn{3}{l}{\emph{SAM-based approaches} }\\
    \hline
    SAM~\cite{kirillov2023segment} &  ViT-L~\cite{dosovitskiy2020image} & 0 & - & - &65.6 & 1.3 \\
    SAM*~\cite{kirillov2023segment} & ViT-L~\cite{dosovitskiy2020image} & 0 & - & - &79.6 & 6.7\\
    Matcher~\cite{qiao2021defrcn}& ViT-L~\cite{dosovitskiy2020image} &1 & 8.0 & 88.9& 23.9 & 22.0 \\
   Crowd-SAM & ViT-L\cite{dosovitskiy2020image} & 10 & 71.4 & 75.1 & 83.9 & 1.7\\
    Crowd-SAM* & ViT-L\cite{dosovitskiy2020image} & 10 & 78.4 & 74.8 & 85.6 & 8.1 \\
    \bottomrule
    \end{tabular}
    }

    \label{tab:main_results}
\end{table}

\textbf{Datasets.} Following \cite{zheng2022progressive}, we adopt CrowdHuman\cite{shao2018crowdhuman} as the benchmark to conduct main experiments and ablation studies. CrowdHuman~\cite{shao2018crowdhuman} collects and annotates images containing crowded persons from the Internet. It contains 15,000, 4,370, and 5,000 images for training, validation, and testing, respectively. We also evaluate our method on CityPersons\cite{zhang2017citypersons} for a realistic urban-scene scenario. Additionally, we utilize OCC-Human~\cite{zhang2019pose2seg}, which is specially reputed for occluded persons. For these pedestrian datasets, we use visible annotations (only including visible areas of an object) for training and evaluation. To validate the extensibility of Crowd-SAM, we further devise a multi-class version of Crowd-SAM by adding a multi-class classifier. We employ 0.1 \% percent of the COCO~\cite{coco}  \emph{trainval} set for training and the COCO \emph{val} set for validation. Besides, we validate our method on a subset with occluded objects on COCO, \ie COCO-OCC~\cite{ke2021deep}, extracted by selecting the images whose objects have a high overlapping ratio.

\textbf{Implementation Details}. We utilize SAM (ViT-L) \cite{kirillov2023segment} and DINOv2 (ViT-L)~\cite{oquab2023dinov2} as the base models for all experiments. In the fine-tuning stage, all their parameters are frozen to avoid over-fitting. Instead of real GT, we use the pseudo masks generated by SAM as $S_{target}$ to supervise the learning of PWD-Net in \cref{eq:iou_loss}. These generated pseudo labels are of high quality as shown in \cref{fig:qualitve_comparison}(c). We randomly pick the points from the pseudo masks as positive training samples and the ones from the background as negative training samples. In the training process, we use Adam\cite{kingma2014adam} with a learning rate of $10^{-5}$, a weight decay of $10^{-4}$, $\beta_{1}=0.9$, and $\beta_{2}=0.99$ for optimization.  We train our module for 2,000 iterations with a batch size of 1, which can be done on a single GTX 3090 Ti GPU in several minutes. For more details, please refer to the Appendix.




\textbf{Evaluation Metrics}. Following \cite{zheng2022progressive}, we use AP with IoU threshold at 0.5, MR$^{-2}$, and Recall as metrics. Generally, a higher AP, Recall, and lower MR$^{-2}$ value indicates better performance.

\subsection{Experimental Results on Pedestrian Detection}
For fair comparison, we re-implement the counterparts\cite{atss,tian2019fcos,zheng2022progressive,dino,qiao2021defrcn} with a $2\times$ schedule using visible annotations in the CrowdHuman~\cite{shao2018crowdhuman} and CityPersons~\cite{zhang2017citypersons} datasets.

\textbf{Main Results.} We compare Crowd-SAM with most related methods including \emph{fully-supervised object detectors}~\cite{tian2019fcos,atss,dino,zheng2022progressive}, \emph{few shot object detectors}~\cite{qiao2021defrcn} and \emph{SAM-based methods}~\cite{liu2024matcher,kirillov2023segment}. Notice that we use visible annotations for which derive different results from those on full box annotations.

\begin{figure}[htb]
    \centering
    \includegraphics[width=1.0\columnwidth]{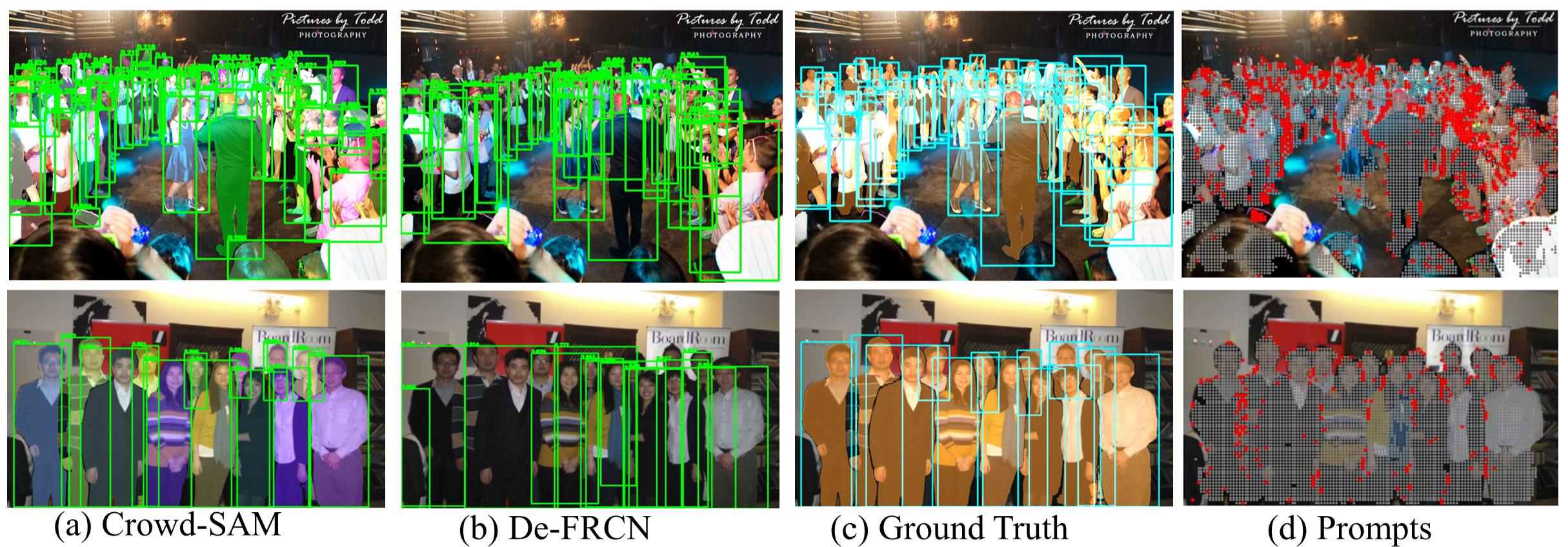}
    \caption{Qualitative comparison between Crowd-SAM (\emph{a}) and De-FRCN (\emph{b}). Crowd-SAM predictions are more accurate especially in the boundaries of persons. We also plot the GT boxes (\emph{\textcolor{blue}{blue rectangles}}) and the generated masks (\emph{\textcolor{orange}{yellow regions}}), which are of high quality (\emph{c}). In (\emph{d}), we plot our prompt filtering results, where preserved prompts (\emph{\textcolor{red}{red points}}) are much fewer than the removed ones (\emph{\textcolor{gray}{gray points}}). Zoom in for a better view.}
    \label{fig:qualitve_comparison}
\end{figure}

As~\cref{tab:main_results} shows, with only 10 labeled images, Crowd-SAM achieves comparable performance with \emph{full-supervised object detectors} whose best result is 86.7\% AP, delivered by DINO\cite{dino}. Particularly, Crowd-SAM outperforms an advanced anchor-free detector FCOS\cite{tian2019fcos} by 2.1\% AP. These results indicate that by using proper adaptation techniques, SAM can reach very competitive performance on complex pedestrian detection datasets like CrowdHuman. On the other side, Crowd-SAM achieves SOTA performance on few-shot detection settings and outperforms all the \emph{few-shot object detectors} by a significant margin. DeFRCN~\cite{qiao2021defrcn} is a well-established few-shot object detector equipped with ResNet-101~\cite{he2016deep} that reports 46.4\% in AP. Notably, our method exceeds it by 32\%, indicating the superiority in the few-shot detection setting. The qualitative comparison between Crowd-SAM and De-FRCN is depicted in~\cref{fig:qualitve_comparison}. For \emph{SAM-based methods}, our Crowd-SAM largely leads the SAM baseline by 6\% (with multi-crop) and 18.3\% (w/o multi-crop) in Recall. Besides, Crowd-SAM is also superior to the other methods such as Matcher~\cite{liu2024matcher}. 

\textbf{Results on OccHuman and CityPersons.} 
To investigate the performance of Crowd-SAM in occluded scenes, we compare it with an advanced counterpart Pose2Seg~\cite{zhang2019pose2seg}, and the results are shown in \cref{tab:comp_occhuman}. It is noteworthy that Pose2Seg is a fully-supervised detector while Crowd-SAM is a few-shot detector. As can be seen from the results, Crowd-SAM leads Pose2Seg by 9.2\% AP which demonstrates its robustness in occluded scenes. Also, we apply our method to a not-so-crowded but more realistic urban dataset, \ie CityPersons~\cite{zhang2017citypersons}, as reported in \cref{tab:comp_cityperson}. Crowd-SAM outperforms TFA~\cite{wang2020frustratingly} by 2.7\% AP, FSCE~\cite{sun2021fsce} by 1.1\% AP, and De-FRCN~\cite{qiao2021defrcn} by 7.8\% AP.  In conclusion, our method remains competitive compared to advanced few-shot object detectors, even though it is not specifically designed for sparse scenes.

These results illustrate that Crowd-SAM well unleashes the power of vision foundation models, \ie SAM and DINO, in all of the crowded, occluded, and urban scenes.

\begin{table}[t]
    \centering
        \caption{Comparative results (\%) on OccHuman \emph{val}, where AP$_M$ and AP$_H$ represent AP in moderate and hard cases according to occlusion ratios, respectively. }
    \scalebox{0.95}{
    \begin{tabular}{lc|ccc}
    \toprule
        Method & Backbone & AP & AP$_M$ & AP$_H$ \\
        \hline
        Mask R-CNN~\cite{mask}& ResNet50-FPN~\cite{he2016deep} & 16.3 & 19.4 & 11.3 \\
        Pose2Seg~\cite{zhang2019pose2seg} & ResNet50-FPN~\cite{he2016deep} & 22.2 & 26.1 & 15.0 \\
        Crowd-SAM &  ViT-L~\cite{dosovitskiy2020image} & \textbf{31.4} & \textbf{26.5} & \textbf{17.7} \\
    \bottomrule
    \end{tabular}
    }

    \label{tab:comp_occhuman}
\end{table}
\begin{table}[t]
    \centering
         \caption{Comparative results (\%) on CityPersons \emph{val}.}
    \scalebox{0.95}{
    \begin{tabular}{lc|cc}
    \toprule
    Method  & \# Shot & AP  & MR$^{-2}$ \\
    \hline
    FCOS\cite{tian2019fcos} & Full & 58.8 &30.0  \\
    ATSS\cite{atss}& Full & 54.1 & 27.8 \\
    Iter-SRCNN\cite{zheng2022progressive} & Full & 57.9& 31.0 \\
    \hline
    TFA\cite{wang2020frustratingly} & 50 & 30.6 &53.8  \\
    FSCE\cite{sun2021fsce} & 50 & 32.2 &  46.5  \\
    De-FRCN\cite{qiao2021defrcn} & 50 &25.5 & 67.1 \\
    \hline
    Crowd-SAM  & 50 & 33.3 & 31.7 \\
    \bottomrule
    \end{tabular}
    }

    \label{tab:comp_cityperson}
\end{table}
\begin{table}[t]
    \centering
        \caption{Comparative results (\%) on COCO \emph{val} and COCO-OCC. }
    \scalebox{0.9}{
    \begin{tabular}{lc|cc|cc}
    \toprule
        Methods & Backbone & 
        \multicolumn{2}{|l|}{COCO-OCC} & \multicolumn{2}{|l}{COCO} \\
        & &AP& AP$_{50}$&  AP& AP$_{50}$\\
        \hline
        Faster R-CNN~\cite{faster} & ResNet50-FPN~\cite{he2016deep} & 29.7 & 50.0 & 33.5 & 53.7 \\
        BCNet~\cite{ke2021deep} & ResNet50-FPN~\cite{he2016deep} & 31.7 & 51.1 & 34.6 & 54.4 \\
        Crowd-SAM &  ViT-L~\cite{dosovitskiy2020image} & 20.6 & 31.5 & 22.0 & 33.7 \\
    \bottomrule
    \end{tabular}
    }

    \label{tab:comp_coco}
\end{table}

\subsection{Experimental Results on Multi-class Object Detection}
To further explore the extensibility in a more popular setting, we devise a multi-class Crowd-SAM. The multi-class version of Crowd-SAM is slightly different by replacing the binary classifier with a multi-class one.  We then validate multi-class Crowd-SAM on COCO~\cite{coco}, a widely adopted common object detection benchmark, and COCO-OCC~\cite{ke2021deep} which is a split of COCO that mainly consists of images with a high occlusion ratio.

We compare our method with two supervised detectors, \ie Faster R-CNN~\cite{faster} and BCNet~\cite{ke2021deep}, and report the results in \cref{tab:comp_coco}. It can be seen that our Crowd-SAM is comparable to the supervised detectors on both datasets and drops only 1.4 AP\% when comparing those of COCO and COCO-OCC. This minor drop indicates that our method is robust to occlusions. 


\begin{table}[htb]
    \centering
    \caption{Ablation results (\%) on the main components in Crowd-SAM. FG location represents the use of the binary classifier on DINOv2 features. The last line represents a SAM baseline with a standard $32 \times 32$ grid as inputs. }
    \scalebox{0.9}{
    \begin{tabularx}{\textwidth}{CCCCCCC}
    \toprule
    FG loc.  & EPS & PWD-Net & Multi-Crop & AP  & MR$^{-2}$ & Recall \\
    \hline
    \checkmark & \checkmark & \checkmark & \checkmark & 78.4 & 74.8 & 85.6 \\
    & \checkmark & \checkmark & \checkmark &  71.0 & 77.9 & 77.1 \\
    \checkmark &  & \checkmark & \checkmark & 77.8 & 71.8 & 83.9 \\
    \checkmark & \checkmark &  & \checkmark & 17.0&  99.6 & 83.1 \\
    \checkmark & \checkmark & \checkmark & & 71.4 & 75.1 & 83.9 \\
      &  &  &  \checkmark &  8.7 & 99.8 & 70.1\\
    \bottomrule
    \end{tabularx}}

    \label{tab:ablation_main}
\end{table}

\subsection{Ablation Studies}
\textbf{Ablation on Modules.} We conduct ablation studies on the key components of Crowd-SAM, including foreground location, EPS, and PWD-Net, to validate their effectiveness. In \cref{tab:ablation_main}, the performance in AP significantly drops by 7.4\% and recall by 8.5\% when FG location is removed, indicating the importance of restricting foreground areas. As for EPS, once it is removed, the AP drops by 0.6\%. We suppose that EPS not only accelerates the sampling process of dense prompts but also helps focus on the difficult part of the image, which conveys ambiguous semantics. We also compare EPS with some other batch iterators in~\cref{tab:ablation_on_EPS}. Besides, we find that PWD-Net is indispensable and when removed, the AP dramatically drops to 17.0\%. Finally, we point out that multi-cropping is a strong trick to enhance the performance which contributes 7.0\% AP to the final performance. Overall, these results prove that all the components are essential. 

\begin{table}[]
    \centering
       \caption{Comparison (\%) of different samplers on CrowdHuman~\cite{shao2018crowdhuman} \emph{val}. Full means using all prompts. OOM represents out-of-memory errors which occur when the GPU memory is all consumed.}
    \scalebox{0.9}{
    \begin{tabularx}{1.0\textwidth}{CC|CC|CC|CC}
    \toprule
          \multicolumn{2}{c}{}  &  \multicolumn{2}{c}{Full} & \multicolumn{2}{c}{Random} &\multicolumn{2}{c}{EPS}  \\
           \hline
           Grid & $K$ & AP & Recall  & AP & Recall & AP  & Recall\\
           \hline
          32 $\times$ 32 & 500  & 57.6& 60.5&   57.6& 60.5 &57.0 & 60.0  \\
          64 $\times$ 64& 500 & 69.4& 73.9&   69.4& 73.9  &69.3 & 73.6  \\
          128 $\times$ 128& 500 &  \multicolumn{2}{c|}{OOM}& 69.7& 74.1  &72.4 & 77.8  \\
          192 $\times$ 192 & 500 & \multicolumn{2}{c|}{OOM} & 69.8 & 73.7   & \textbf{73.2} & \textbf{78.2}  \\
          256 $\times$ 256 & 500 & \multicolumn{2}{c|}{OOM} & 68.9 & 73.0  & 72.3 & 78.0  \\

    \bottomrule
    \end{tabularx}
    }
 
    \label{tab:ablation_on_EPS}
\end{table}

\textbf{Ablation on EPS}. We conduct a comprehensive study on EPS by comparing it with several variants. We forward each image only once to avoid the extra latency caused by multi-crop. We use a default sampler that iterates through all training samples and a random iterator with a halting threshold $K$ as counter-parts. As reported in~\cref{tab:ablation_on_EPS}, the AP and Recall increase with the grid size for all three samplers. However, the default sampler suffers an out-of-memory error when the grid size reaches 128, preventing it from being adopted in this setting. As for the random sampler, its performance is constrained by $K$ and a grid size larger than 64 only leads to limited improvement, \eg 0.1\% AP. On the contrary, our EPS benefits from a much larger grid size and achieves a better result.

\begin{table}[]
      \caption{Ablation results (\%) on the design of PWD-Net. ${\cal M}$, ${\cal U}$, and ${\cal O}$ represent the mask token, IoU token, and semantic token, respectively. For the IoU head, $F$ means freezing the original IoU head and training a parallel one, and $T$ indicates tuning the original IoU head. }
    \centering
    \scalebox{0.9}{
    \begin{tabular}{p{0.7cm}p{0.7cm}p{0.7cm}c|cc}
    \toprule
    ${\cal M}$  & ${\cal U}$ & ${\cal O}$ &  IoU head  & AP  & MR$^{-2} $\\
    \hline
    \checkmark & \checkmark & \checkmark & $F$ & 78.4 & 74.8 \\
    \checkmark & \checkmark & \checkmark &  $T$ & 75.6({-2.8}) & 80.4({+5.6}) \\
    \checkmark & \checkmark &  & $F$ & 77.3({-1.1}) & 73.5({-1.3}) \\
    \checkmark &  & \checkmark & $F$ & 76.3({-2.1}) & 74.8({-0.0}) \\
     & \checkmark & \checkmark &  $F$ & 38.4({-40.0}) & 95.8({+21.0}) \\
    \bottomrule
    \end{tabular}
    }
  
    \label{tab:ablation_pwdnet}
\end{table}

\textbf{Ablation on PWD-Net.} We compare PWD-Net to the variants that replace some tokens with a full-zero placeholder. We also consider two designs, directly tuning the IoU head or learning a parallel IoU head. As shown in ~\cref{tab:ablation_pwdnet}, all the three tokens, \ie Mask Token ${\cal M}$, IoU Token ${\cal U}$, and Semantic Token ${\cal O}$, contribute to the final result. Particularly, once ${\cal M}$ is removed, the AP drops by 40.0\%, which is a catastrophic decline. This degradation suggests that the mask token contains the shape-aware feature that is essential for the part-whole discrimination task. Notably, the AP drops by 2.8\% when we tune the pre-trained IoU Head, suggesting that it is prone to overfit the few labeled images. By freezing the IoU head of SAM, PWD-Net can benefit more from the shape-aware knowledge learned from massive segmentation data.

\section{Conclusion}
This paper proposes Crowd-SAM, a SAM-based framework, for object detection and segmentation in crowded scenes, designed to streamline the annotation process. 
For each image, Crowd-SAM generates dense prompts for high recall and uses EPS to prune redundant prompts. To achieve accurate detection in occlusion cases, Crowd-SAM employs PWD-Net which leverages several informative tokens to select the masks that best fit. Combined with the proposed modules, Crowd-SAM achieves 78.4\% AP on CrowdHuman, comparable to full-supervised detectors, validating that object detection in crowded scenes can benefit from foundation models like SAM with data efficiency.
\section*
{Acknowledgements}
{This work is partly supported by the National Natural Science Foundation of China (No. 62022011), the Research Program of State Key Laboratory of Software Development Environment, and the Fundamental Research Funds for the Central Universities.}
%
%
\bibliographystyle{splncs04}
\bibliography{main}

\begin{thebibliography}{10}
\providecommand{\url}[1]{\texttt{#1}}
\providecommand{\urlprefix}{URL }
\providecommand{\doi}[1]{https://doi.org/#1}

\bibitem{bar2022detreg}
Bar, A., Wang, X., Kantorov, V., Reed, C.J., Herzig, R., Chechik, G., Rohrbach, A., Darrell, T., Globerson, A.: Detreg: Unsupervised pretraining with region priors for object detection. In: IEEE Conf. Comput. Vis. Pattern Recog. pp. 14605--14615 (2022)

\bibitem{detr}
Carion, N., Massa, F., Synnaeve, G., Usunier, N., Kirillov, A., Zagoruyko, S.: End-to-end object detection with transformers. In: Eur. Conf. Comput. Vis. pp. 213--229. Springer (2020)

\bibitem{caron2021emerging}
Caron, M., Touvron, H., Misra, I., J{\'e}gou, H., Mairal, J., Bojanowski, P., Joulin, A.: Emerging properties in self-supervised vision transformers. In: Int. Conf. Comput. Vis. pp. 9650--9660 (2021)

\bibitem{chen2024rsprompter}
Chen, K., Liu, C., Chen, H., Zhang, H., Li, W., Zou, Z., Shi, Z.: Rsprompter: Learning to prompt for remote sensing instance segmentation based on visual foundation model. IEEE Transactions on Geoscience and Remote Sensing  (2024)

\bibitem{chi2020pedhunter}
Chi, C., Zhang, S., Xing, J., Lei, Z., Li, S.Z., Zou, X.: Pedhunter: Occlusion robust pedestrian detector in crowded scenes. In: AAAI. vol.~34, pp. 10639--10646 (2020)

\bibitem{dai2021up}
Dai, Z., Cai, B., Lin, Y., Chen, J.: Up-detr: Unsupervised pre-training for object detection with transformers. In: IEEE Conf. Comput. Vis. Pattern Recog. pp. 1601--1610 (2021)

\bibitem{dosovitskiy2020image}
Dosovitskiy, A., Beyer, L., Kolesnikov, A., Weissenborn, D., Zhai, X., Unterthiner, T., Dehghani, M., Minderer, M., Heigold, G., Gelly, S., et~al.: An image is worth 16x16 words: Transformers for image recognition at scale. In: Int. Conf. Learn. Represent. (2021)

\bibitem{gao2023selecting}
Gao, F., Leng, J., Gan, J., Gao, X.: Selecting learnable training samples is all detrs need in crowded pedestrian detection. In: ACM Int. Conf. Multimedia. pp. 2714--2722 (2023)

\bibitem{girshick2015fast}
Girshick, R.: Fast r-cnn. In: Int. Conf. Comput. Vis. pp. 1440--1448 (2015)

\bibitem{girshick2014rich}
Girshick, R., Donahue, J., Darrell, T., Malik, J.: Rich feature hierarchies for accurate object detection and semantic segmentation. In: IEEE Conf. Comput. Vis. Pattern Recog. pp. 580--587 (2014)

\bibitem{grill2020bootstrap}
Grill, J.B., Strub, F., Altch{\'e}, F., Tallec, C., Richemond, P., Buchatskaya, E., Doersch, C., Avila~Pires, B., Guo, Z., Gheshlaghi~Azar, M., et~al.: Bootstrap your own latent-a new approach to self-supervised learning. Adv. Neural Inform. Process. Syst.  \textbf{33},  21271--21284 (2020)

\bibitem{gui2024remote}
Gui, S., Song, S., Qin, R., Tang, Y.: Remote sensing object detection in the deep learning era—a review. Remote Sensing  \textbf{16}(2), ~327 (2024)

\bibitem{mask}
He, K., Gkioxari, G., Doll{\'a}r, P., Girshick, R.: Mask r-cnn. In: Int. Conf. Comput. Vis. pp. 2961--2969 (2017)

\bibitem{he2016deep}
He, K., Zhang, X., Ren, S., Sun, J.: Deep residual learning for image recognition. In: IEEE Conf. Comput. Vis. Pattern Recog. pp. 770--778 (2016)

\bibitem{hoiem2009pascal}
Hoiem, D., Divvala, S.K., Hays, J.H.: Pascal voc 2008 challenge. World Literature Today  \textbf{24}(1), ~1--4 (2009)

\bibitem{kang2019few}
Kang, B., Liu, Z., Wang, X., Yu, F., Feng, J., Darrell, T.: Few-shot object detection via feature reweighting. In: Int. Conf. Comput. Vis. pp. 8420--8429 (2019)

\bibitem{ke2021deep}
Ke, L., Tai, Y.W., Tang, C.K.: Deep occlusion-aware instance segmentation with overlapping bilayers. In: IEEE Conf. Comput. Vis. Pattern Recog. pp. 4019--4028 (2021)

\bibitem{hqsam}
Ke, L., Ye, M., Danelljan, M., Tai, Y.W., Tang, C.K., Yu, F., et~al.: Segment anything in high quality. In: Adv. Neural Inform. Process. Syst. vol.~36 (2024)

\bibitem{kingma2014adam}
Kingma, D.P., Ba, J.: Adam: A method for stochastic optimization. In: Int. Conf. Learn. Represent. (2015)

\bibitem{kirillov2023segment}
Kirillov, A., Mintun, E., Ravi, N., Mao, H., Rolland, C., Gustafson, L., Xiao, T., Whitehead, S., Berg, A.C., Lo, W.Y., et~al.: Segment anything. In: Int. Conf. Comput. Vis. pp. 4015--4026 (2023)

\bibitem{Li_2023_ICCV}
Li, M., Wu, J., Wang, X., Chen, C., Qin, J., Xiao, X., Wang, R., Zheng, M., Pan, X.: Aligndet: Aligning pre-training and fine-tuning in object detection. In: Int. Conf. Comput. Vis. pp. 6866--6876 (2023)

\bibitem{lin2020detr}
Lin, M., Li, C., Bu, X., Sun, M., Lin, C., Yan, J., Ouyang, W., Deng, Z.: Detr for crowd pedestrian detection. arXiv preprint arXiv:2012.06785  (2020)

\bibitem{lin2017focal}
Lin, T.Y., Goyal, P., Girshick, R., He, K., Doll{\'a}r, P.: Focal loss for dense object detection. In: Int. Conf. Comput. Vis. pp. 2980--2988 (2017)

\bibitem{coco}
Lin, T.Y., Maire, M., Belongie, S., Hays, J., Perona, P., Ramanan, D., Doll{\'a}r, P., Zitnick, C.L.: Microsoft coco: Common objects in context. In: Eur. Conf. Comput. Vis. pp. 2906--2917 (2014)

\bibitem{Liu_2019_CVPR}
Liu, S., Huang, D., Wang, Y.: Adaptive nms: Refining pedestrian detection in a crowd. In: IEEE Conf. Comput. Vis. Pattern Recog. (June 2019)

\bibitem{liu2020self}
Liu, S., Li, Z., Sun, J.: Self-emd: Self-supervised object detection without imagenet. arXiv preprint arXiv:2011.13677  (2020)

\bibitem{liu2016ssd}
Liu, W., Anguelov, D., Erhan, D., Szegedy, C., Reed, S., Fu, C.Y., Berg, A.C.: Ssd: Single shot multibox detector. In: Eur. Conf. Comput. Vis. pp. 21--37. Springer (2016)

\bibitem{liu2024matcher}
Liu, Y., Zhu, M., Li, H., Chen, H., Wang, X., Shen, C.: Matcher: Segment anything with one shot using all-purpose feature matching. In: Int. Conf. Learn. Represent. (2024)

\bibitem{liu2021unbiased}
Liu, Y.C., Ma, C.Y., He, Z., Kuo, C.W., Chen, K., Zhang, P., Wu, B., Kira, Z., Vajda, P.: Unbiased teacher for semi-supervised object detection. In: Int. Conf. Learn. Represent. (2021)

\bibitem{Liu_2022_CVPR}
Liu, Y.C., Ma, C.Y., Kira, Z.: Unbiased teacher v2: Semi-supervised object detection for anchor-free and anchor-based detectors. In: IEEE Conf. Comput. Vis. Pattern Recog. pp. 9819--9828 (2022)

\bibitem{ma2024segment}
Ma, J., He, Y., Li, F., Han, L., You, C., Wang, B.: Segment anything in medical images. Nature Communications  \textbf{15}(1), ~654 (2024)

\bibitem{mao2017can}
Mao, J., Xiao, T., Jiang, Y., Cao, Z.: What can help pedestrian detection? In: IEEE Conf. Comput. Vis. Pattern Recog. pp. 3127--3136 (2017)

\bibitem{oquab2023dinov2}
Oquab, M., Darcet, T., Moutakanni, T., Vo, H., Szafraniec, M., Khalidov, V., Fernandez, P., Haziza, D., Massa, F., El-Nouby, A., et~al.: Dinov2: Learning robust visual features without supervision. Trans. Mach. Learn Res.  (2024)

\bibitem{qiao2021defrcn}
Qiao, L., Zhao, Y., Li, Z., Qiu, X., Wu, J., Zhang, C.: Defrcn: Decoupled faster r-cnn for few-shot object detection. In: Int. Conf. Comput. Vis. pp. 8681--8690 (2021)

\bibitem{redmon2016you}
Redmon, J., Divvala, S., Girshick, R., Farhadi, A.: You only look once: Unified, real-time object detection. In: IEEE Conf. Comput. Vis. Pattern Recog. pp. 779--788 (2016)

\bibitem{faster}
Ren, S., He, K., Girshick, R., Sun, J.: Faster r-cnn: Towards real-time object detection with region proposal networks. In: Adv. Neural Inform. Process. Syst. vol.~28 (2015)

\bibitem{shao2018crowdhuman}
Shao, S., Zhao, Z., Li, B., Xiao, T., Yu, G., Zhang, X., Sun, J.: Crowdhuman: A benchmark for detecting human in a crowd. arXiv preprint arXiv:1805.00123  (2018)

\bibitem{su2012crowdsourcing}
Su, H., Deng, J., Fei-Fei, L.: Crowdsourcing annotations for visual object detection. In: Workshops at the twenty-sixth AAAI conference on artificial intelligence (2012)

\bibitem{sun2021fsce}
Sun, B., Li, B., Cai, S., Yuan, Y., Zhang, C.: Fsce: Few-shot object detection via contrastive proposal encoding. In: IEEE Conf. Comput. Vis. Pattern Recog. pp. 7352--7362 (2021)

\bibitem{sparsercnn}
Sun, P., Zhang, R., Jiang, Y., Kong, T., Xu, C., Zhan, W., Tomizuka, M., Li, L., Yuan, Z., Wang, C., et~al.: Sparse r-cnn: End-to-end object detection with learnable proposals. In: IEEE Conf. Comput. Vis. Pattern Recog. pp. 14454--14463 (2021)

\bibitem{tang2017multiple}
Tang, P., Wang, X., Bai, X., Liu, W.: Multiple instance detection network with online instance classifier refinement. In: IEEE Conf. Comput. Vis. Pattern Recog. pp. 2843--2851 (2017)

\bibitem{tang2021humble}
Tang, Y., Chen, W., Luo, Y., Zhang, Y.: Humble teachers teach better students for semi-supervised object detection. In: IEEE Conf. Comput. Vis. Pattern Recog. pp. 3132--3141 (2021)

\bibitem{tian2019fcos}
Tian, Z., Shen, C., Chen, H., He, T.: Fcos: Fully convolutional one-stage object detection. In: Int. Conf. Comput. Vis. pp. 9627--9636 (2019)

\bibitem{wang2020frustratingly}
Wang, X., Huang, T.E., Darrell, T., Gonzalez, J.E., Yu, F.: Frustratingly simple few-shot object detection. Int. Conf. Mach. Learn.  (2020)

\bibitem{wang2018repulsion}
Wang, X., Xiao, T., Jiang, Y., Shao, S., Sun, J., Shen, C.: Repulsion loss: Detecting pedestrians in a crowd. In: IEEE Conf. Comput. Vis. Pattern Recog. pp. 7774--7783 (2018)

\bibitem{wei2023semantic}
Wei, Z., Chen, P., Yu, X., Li, G., Jiao, J., Han, Z.: Semantic-aware sam for point-prompted instance segmentation. In: IEEE Conf. Comput. Vis. Pattern Recog. pp. 3585--3594 (2024)

\bibitem{wu2023medical}
Wu, J., Fu, R., Fang, H., Liu, Y., Wang, Z., Xu, Y., Jin, Y., Arbel, T.: Medical sam adapter: Adapting segment anything model for medical image segmentation. arXiv preprint arXiv:2304.12620  (2023)

\bibitem{xie2021detco}
Xie, E., Ding, J., Wang, W., Zhan, X., Xu, H., Sun, P., Li, Z., Luo, P.: Detco: Unsupervised contrastive learning for object detection. In: Int. Conf. Comput. Vis. pp. 8392--8401 (2021)

\bibitem{xu2021end}
Xu, M., Zhang, Z., Hu, H., Wang, J., Wang, L., Wei, F., Bai, X., Liu, Z.: End-to-end semi-supervised object detection with soft teacher. In: Int. Conf. Comput. Vis. pp. 3060--3069 (2021)

\bibitem{zhao2023fast}
Xu, Z., Wenchao, D., Yongqi, A., Yinglong, D., Tao, Y., Min, L., Ming, T., Jinqiao, W.: Fast segment anything. arXiv preprint arXiv:2306.12156  (2023)

\bibitem{yan2019meta}
Yan, X., Chen, Z., Xu, A., Wang, X., Liang, X., Lin, L.: Meta r-cnn: Towards general solver for instance-level low-shot learning. In: Int. Conf. Comput. Vis. pp. 9577--9586 (2019)

\bibitem{ye2024sam}
Ye, Z., Lovell, L., Faramarzi, A., Ninic, J.: Sam-based instance segmentation models for the automation of masonry crack detection. arXiv preprint arXiv:2401.15266  (2024)

\bibitem{Zeng_2019_ICCV}
Zeng, Z., Liu, B., Fu, J., Chao, H., Zhang, L.: Wsod2: Learning bottom-up and top-down objectness distillation for weakly-supervised object detection. In: Int. Conf. Comput. Vis. (2019)

\bibitem{zhang2023faster}
Zhang, C., Han, D., Qiao, Y., Kim, J.U., Bae, S.H., Lee, S., Hong, C.S.: Faster segment anything: Towards lightweight sam for mobile applications. arXiv preprint arXiv:2306.14289  (2023)

\bibitem{dino}
Zhang, H., Li, F., Liu, S., Zhang, L., Su, H., Zhu, J., Ni, L., Shum, H.Y.: {DINO}: {DETR} with improved denoising anchor boxes for end-to-end object detection. In: Int. Conf. Learn. Represent. (2023)

\bibitem{zhang2019double}
Zhang, K., Xiong, F., Sun, P., Hu, L., Li, B., Yu, G.: Double anchor r-cnn for human detection in a crowd. arXiv preprint arXiv:1909.09998  (2019)

\bibitem{zhang2024personalize}
Zhang, R., Jiang, Z., Guo, Z., Yan, S., Pan, J., Dong, H., Qiao, Y., Gao, P., Li, H.: Personalize segment anything model with one shot. In: Int. Conf. Learn. Represent. (2024)

\bibitem{zhang2017citypersons}
Zhang, S., Benenson, R., Schiele, B.: Citypersons: A diverse dataset for pedestrian detection. In: IEEE Conf. Comput. Vis. Pattern Recog. pp. 3213--3221 (2017)

\bibitem{atss}
Zhang, S., Chi, C., Yao, Y., Lei, Z., Li, S.Z.: Bridging the gap between anchor-based and anchor-free detection via adaptive training sample selection. In: IEEE Conf. Comput. Vis. Pattern Recog. pp. 9759--9768 (2020)

\bibitem{zhang2018occlusion}
Zhang, S., Wen, L., Bian, X., Lei, Z., Li, S.Z.: Occlusion-aware r-cnn: Detecting pedestrians in a crowd. In: Eur. Conf. Comput. Vis. pp. 637--653 (2018)

\bibitem{zhang2019pose2seg}
Zhang, S.H., Li, R., Dong, X., Rosin, P., Cai, Z., Han, X., Yang, D., Huang, H., Hu, S.M.: Pose2seg: Detection free human instance segmentation. In: IEEE Conf. Comput. Vis. Pattern Recog. pp. 889--898 (2019)

\bibitem{zheng2022progressive}
Zheng, A., Zhang, Y., Zhang, X., Qi, X., Sun, J.: Progressive end-to-end object detection in crowded scenes. In: IEEE Conf. Comput. Vis. Pattern Recog. pp. 857--866 (2022)

\bibitem{deformabledetr}
Zhu, X., Su, W., Lu, L., Li, B., Wang, X., Dai, J.: Deformable detr: Deformable transformers for end-to-end object detection. In: Int. Conf. Learn. Represent. (2021)

\end{thebibliography}
\end{document}